\documentclass{article}
\usepackage[preprint]{neurips_2020}
\usepackage{rotating}
\usepackage{amsmath}
\usepackage{authblk}
\usepackage[utf8]{inputenc} 
\usepackage[T1]{fontenc}    
\usepackage{hyperref}       
\usepackage{url}            
\usepackage{booktabs}       
\usepackage{amsfonts}       
\usepackage{nicefrac}       
\usepackage{microtype}      
\usepackage{xcolor}         

\title{Comparing Machine Learning Techniques for Alfalfa  Biomass Yield Prediction}

\author[ , 1]{Jonathan M. Vance\thanks{jmvance@uga.edu}}
\author[ , 2]{Christopher D. Whitmire\thanks{cwhitmire@berry.edu}}
\author[ , 2]{Hend K. Rasheed\thanks{hend.rasheed25@uga.edu}}
\author[ , 1]{Christian Adkins\thanks{caadkins@uga.edu}}
\author[ , 3]{Ali Missaoui\thanks{cssamm@uga.edu}}
\author[ , 1, 2]{Khaled M. Rasheed\thanks{khaled@uga.edu (Contact Author)}}
\author[  , 2]{Frederick W. Maier\thanks{fmaier@uga.edu}}
\affil[1]{Department of Computer Science, University of Georgia, 415 Boyd Graduate Studies, 200 D. W. Brooks Drive, Athens, GA 30602, USA}
\affil[2]{Institute for Artificial Intelligence, University of Georgia, 515 Boyd Graduate Studies, 200 D. W. Brooks Drive, Athens, GA 30602, USA}
\affil[3]{Department of Crop and Soil Sciences, Institute of Plant Breeding Genetics and Genomics,
University of Georgia, 4317 Miller Plant Science, Athens, GA 30602, USA}

\begin{document}
  
  \maketitle{}
  
  \section*{}
  Keywords: alfalfa; feature selection; biomass; precision agriculture; yield prediction; climate change
  \section*{Submission Type: Regular Research Paper}


\begin{abstract}
The alfalfa crop is globally important as livestock feed, so highly efficient planting and harvesting could benefit many industries, especially as the global climate changes and traditional methods become less accurate. Recent work using machine learning (ML) to predict yields for alfalfa and other crops has shown promise. Previous efforts used remote sensing, weather, planting, and soil data to train machine learning models for yield prediction. However, while remote sensing works well, the models require large amounts of data and cannot make predictions until the harvesting season begins. Using weather and planting data from alfalfa variety trials in Kentucky and Georgia, our previous work compared feature selection techniques to find the best technique and best feature set. In this work, we trained a variety of machine learning models, using cross validation for hyperparameter optimization, to predict biomass yields, and we showed better accuracy than similar work that employed more complex techniques. Our best individual model was a random forest with a mean absolute error of 0.081 tons/acre and R{$^2$} of 0.941. Next, we expanded this dataset to include Wisconsin and Mississippi, and we repeated our experiments, obtaining a higher best R{$^2$} of 0.982 with a regression tree. We then isolated our testing datasets by state to explore this problem's eligibility for domain adaptation (DA), as we trained on multiple source states and tested on one target state. This Trivial DA (TDA) approach leaves plenty of room for improvement through exploring more complex DA techniques in forthcoming work.
\end{abstract}



\section{Introduction}%
\label{introduction}


With the intent of directing world leaders toward solving some of the world’s biggest problems, the United Nations recently developed a collection of 17 goals and 169 targets. The hope is that the world will reach these goals by the year 2030 {\citep{UN}}. However, it is the opinion of the Copenhagen Consensus Center (CCC), a think tank, that appropriately prioritizing these goals will increase the likelihood that the world will reach them {\citep{CCC}}.  The CCC has performed a cost-benefit analysis on all these targets and ranked them accordingly. One of their findings was that increasing research and development into maximizing global crop yields would be one of the most cost-effective ways of achieving the UN’s goals {\citep{rosegrant2018returns}}.  Specifically, every \$1 spent on this kind of R\&D would result in \$34 worth of benefit worldwide. \citep{lomborg2015nobel} 
One possible way to increase yields is to improve agricultural planning, which would help ensure that there are sufficient yields of key crops. At the start of every season, agricultural planners need to estimate the yields of different agricultural plans \citep{frausto2009new}. Often, farmers rely on their own personal experiences of history to predict what their yields will be, but this can result in limited accuracy \citep{russ2009data}. Given that crop yields vary spatially and temporally, and are sensitive to varying conditions like weather, other prediction methods should be investigated.
The USDA, with its National Agricultural Statistics Service branch, makes monthly forecasts of crop yields in the United States. It does this by conducting two surveys, a farm operator survey and an objective survey. The farm operator survey is done by calling farmers at random and asking them what they think their predicted yield for the next month will be. The objective survey involves an investigator going out and surveying random fields and recording data on the output of those fields. The findings of these surveys are compared to previous historical data to confirm that the findings are consistent with previous harvests with similar conditions. The final predicted yields then come from the results of these surveys \citep{NationalAgriculturalStatisticsCenter}\citep{johnson2014assessment}. The findings of this methodology, when compared to the ground truth, have had very low errors \citep{you2017deep}\citep{NationalAgriculturalStatisticsCenter}. However, this process is very resource intensive. The farm operator survey is done primarily over the phone, and the objective survey requires measurements to be taken in person at hundreds of farms every month \citep{NationalAgriculturalStatisticsCenter}\citep{johnson2014assessment}. 
An alternative approach is to use remote sensing (RS) data. RS techniques use images achieved primarily from aircraft of satellites, and these images record spectral, spatial, and temporal information \citep{chlingaryan2018machine}. Mathematical operations can be performed on these images to form vegetation indices (VIs), which can be used as inputs into machine learning algorithms \citep{xue2017significant}. Recent work has been done to use VIs to predict crop yield. You et al., had great success at predicting county level soybean yield in the United States using remote sensing data as input for a convolutional neural network (CNN) and a long short-term memory (LSTM) model, both with a Gaussian Process component \citep{you2017deep}. Panda, Ames, \& Panigrahi used several different VIs as an input to a neural network (NN) to predict corn yield \citep{panda2010application}. Johnson did something similar but used regression trees to predict both corn and soybean yield \citep{johnson2014assessment}. However, despite these successes, there are difficulties with building machine learning models based on remote sensing data.  This is because using remote sensing data depends on the processing of large amounts of data across different platforms \citep{chlingaryan2018machine}. These models also cannot make a prediction unless there are images available for input, which means that this model cannot begin making predictions until the growing season has started \citep{cunha2018scalable}. Xue and Su also compared over one hundred different vegetation indices and found that no VI is universally better than the others. Each is more suitable to certain situations, and each has their own limitations \citep{xue2017significant}. This means that it may be difficult to know the optimal VI for each case.
Weather, spatial, and soil features have also been used to train machine learning models to predict crop yield \citep{gonzalez2014predictive}\citep{ayoubi2011comparing}\citep{jeong2016random}. These kinds of data also require less processing than remote sensing data and can be used to make predictions before the season starts. They also have the potential to use weather forecasting results to make predictions before the season begins, making it more convenient for planning purposes than using remote sensing data. Similarly, the current paper uses weather and planting data to develop a variety of machine learning models and compares the results.

The current work builds further on the previous work by Whitmire et al. by comparing biomass yield prediction experiment results among various ML models, as we explore elements of domain adaptation (DA) to further improve our application pipeline \citep{whitmire2021using}. We used data from different source regions to train models which then predict crop yields in a specific target region. To this end, we achieve three major contributions:

\begin{itemize}
    \item Provide results from ML experiments comparing various models, showing our weather-based ML approach to be promising
    \item Broadly expand the alfalfa ML training dataset, adding curated crop yield and weather data from three additional states
    \item Discover positive results for seemingly disparate sources and targets, such as source: Wisconsin, target: Georgia
\end{itemize}

In Section II, we discuss related research and topics relevant to this project. Section III provides details about our methods as well as the collection and characteristics of the data. Experiment setups and results are discussed in Section IV, Section V contains conclusions, and Section VI outlines directions of future work. 




\section{Related Work}%
\label{related}
\subsection{Background}
This project began as C. Whitmire'’s master'’s thesis, with help from H. Rasheed, under the direction of K. Rasheed et al. at the University of Georgia~\citep{whitmire2019machine}. They aggregated 771 records of alfalfa biomass yield data from Georgia and Kentucky, and they correlated this with weather data from the same time periods; specific weather features include the same ones in the current work, harvested from the NOAA as well as other sources. By experimenting with feature selection techniques, they found that the most salient features were Julian day of harvest, temperature, number of days since the crop was sown, and cumulative rainfall and solar radiation since last harvest \citep{whitmire2021using}. They split the data into train, test, and validation sets to train a neural network (NN), random forest (RF), regression tree (RT), Bayesian Ridge Regression, linear regression, k-nearest neighbors (KNN), and support vector machine (SVM) models to predict yields in the test set. Using k-fold cross-validation, they performed grid searches to determine which model configurations achieved the highest accuracies \citep{whitmire2021using}. Our results approach the accuracy of similar work by You et al. at Stanford University~\citep{you2017deep}, even though You’s datasets are much larger and collected by more sophisticated methods involving remote sensors, which enables the application of much deeper and more complex neural networks. Overall, our results were promising enough to motivate us to explore domain adaptation (DA), where models are trained using one mathematically augmented source dataset to predict a distinct target dataset. We explore DA by attempting a trivial DA (TDA) approach, where we train on source data from our original dataset plus three new U.S. States, and we set the target dataset as GA. 

As the core problem with successfully predicting crop yields using machine learning is scarcity of data, much of the related works focus on techniques designed to accumulate larger datasets. Like You, Xue and Su at Northwest A\&F University in China highlight the benefits of remote sensor (RS) technology, such as image data collected by unmanned aerial vehicles (AEVs)~\citep{xue2017significant}. Xue’s work reviews over 100 vegetation indices (VIs), which are assorted algorithms with no mathematical unifying basis that are customized to measure important characteristics used in crop prediction and related applications. 

Our work most closely resembles research applying deep learning (DL) to yield forecasting soybean and maize by Oliveira et al. at IBM Research~\citep{oliveira2018scalable}. Like ours, that work applies a very simple method of manual yield data collection instead of developing VIs or using RS technology. On the other hand, their work forecasts future yields, while the current work only predicts known yields. Oliveira’s abundance of over 50,000 datapoints opens that work to the use of a deep neural network (DNN), which we would like to try in future work. However, the current work achieves better R{$^2$} scores that fall comfortably in the 0.8 to 0.9+ range, while Oliveira reports R{$^2$} scores ranging from .55 to .75~\citep{oliveira2018scalable,whitmire2019machine}. 

\subsection{Transfer Learning and Domain Adaptation}

Transfer learning is an area of research focused on improving learning in a new task through the transfer of knowledge from a related task that has already been learned. This is in contrast with most machine learning algorithms which are designed to address single tasks~{\citep{torrey2010transfer}}. While useful in many research problems, transfer learning as a whole does not directly apply to this work. The task of predicting alfalfa yield remains the same; what changes is the region that data comes from. These different regions may have significant differences in the underlying distributions of feature values, with varying temperature ranges and standard levels of rainfall for example. 

Thus, a subcategory of transfer learning known as domain adaptation (DA) is more potentially applicable to our research. DA uses data from a source with one distribution to predict target values in a test domain with a different distribution~{\citep{jiang2008literature}}. A large amount of research has utilized DA in the area of sequence labeling~{\citep{daume2009frustratingly}}. In the context of agriculture, image classification is a popular technique for studying aspects such as crop yield~{\citep{bellocchio2020combining}}. But such methods are very expensive and time consuming, are not feasible in all locations, and do not benefit from existing data collection that has been carried out over the years. Our work aims to incorporate DA in order to coalesce such widespread numerical data and achieve meaningful prediction accuracy without relying on expensive image collection and analysis techniques. 
  

\section{Approach}%
\label{approach}
\subsection{ML Approach}
We used the Python programming language throughout this research \citep{Python}. Specifically, Python as provided within the Anaconda environment was used \citep{Anaconda}. The following packages were used: Pandas for data cleaning and preparation \citep{mckinney2010data}, matplotlib \citep{hunter2007matplotlib} and seaborn \citep{michael_waskom_2022_6609266} for visualizations, sci-kit learn to make and evaluate the machine learning models \citep{pedregosa2011scikit}, and finally, numpy for general mathematical operations \citep{oliphant2006guide}\citep{van2011numpy}. 

The features used in training our machine learning models were the Julian day of the harvest, the number of days between the harvest and the sown date of the crop, the cumulative solar radiation since the previous harvest, temperature, and the cumulative rainfall since the last harvest. The cumulative solar radiation and rainfall values were found by summing daily values.
All the data sources for the previous and current work are presented in Appendix A. Alfalfa harvest data was obtained from variety trials conducted by the University of Georgia (UGA) and University of Kentucky (UKY) for our first round of experiments. We added data from Pennsylvania (PA), Wisconsin (WI), and Mississippi (MS) for the second round. These reports detail the yield in tons per acre of multiple varieties of alfalfa. UGA’s data came from Athens and Tifton, GA from the years 2008 to 2010 where harvests occur from April to December. UKY’s trials occur in Lexington, KY ranging from 2013 to 2018 and contain data from the months of May through September. These variety trials reported multiple cut dates per year.

We carefully curated daily weather data for each location. Data for Tifton and Watkinsville, which is about 13 miles from Athens, GA, was retrieved from the Georgia Automated environmental network. Similar data was found for Versailles, which is nearby Lexington, KY, from the National Oceanic and Atmospheric Administration (NOAA).  

Also, all the data points that had harvest dates with the same year as the sown date were filtered out. Similarly, the first harvest of every season was removed because the amount of time since the previous harvest would be much larger for this harvest relative to subsequent harvests. After this cleaning process, we ended up with 770 datapoints total. 

Before training the models, all of the features were normalized according to the formula 
\begin{align*}
  x_{new} &= \frac{x_{old} - x_{mean}}{x_{SDev}}\\
\end{align*}
where \begin{equation*}x_{old}\end{equation*} was the original value of the feature, \begin{equation*}x_{mean}\end{equation*} is the average value of the features, and \begin{equation*}x_{SDev}\end{equation*} is the standard deviation of the values for that feature.
Before training the models in our previous work, the data was shuffled and split into ten folds to be used for 10-fold cross validation. For each fold, a machine learning model was initialized. This means that ten models were made for each method, one model for each fold. Then, within this outer fold, a grid search (Appendix B) with 5-fold cross validation was done to find the hyperparameters for the model that most minimized the mean absolute error. Once the hyperparameters were found, the machine learning model was trained on the training set and was evaluated against the testing set. We calculated the mean absolute error (MAE), R value, and R{$^2$} value. The average errors, percent error, R, and R{$^2$} value over the ten iterations was found and recorded, and the results of the best model were also recorded along with their standard deviations\citep{whitmire2019machine}. 
This process was done to train and evaluate the following methods: regression tree, random forest regression, k-nearest neighbors, support vector machines, neural networks, Bayesian Ridge regression, and linear regression. Once the results for each method were obtained, an unpaired two-tailed t test was used to find the p-value between the average R{$^2$} values of each method\citep{whitmire2019machine}. As shown in Table~\ref{tbl:whitmireResults}, we achieved our highest R{$^2$} score of 0.941 with our Random Forest model. 

We collected data from Mississippi State University (MSU), Penn-State University, and the University of Wisconsin-Madison (UW-Madison). Our MSU data come from Starkville, Poplarville, Holly Springs, and Newton, MS between March and December from 2012 to 2016. Our Penn-State data come from Rock Springs and Landisville, PA annually from 2008 to 2019. Our UW-Madison data come from Lancaster, Marshfield, and Arlington, WS from 2011 to 2016. Some variety trials report annual biomass yields, while others report multiple cuts per year, and all report sown dates. Therefore, we decided to annualize all our data for the current approach, since the reverse is not feasible. We also referred to the NOAA for our PA, WI, and MS weather data, matching to the exact city where possible, and sometimes to nearby localities with more complete data, as in the previous approach. While we made every effort to compile the most accurate data possible for this work, there were cases where we had to make assumptions about some missing weather data from NOAA to compile our dataset.

We selected the following ML models implemented in SciKit-Learn: random forest (RF), k-nearest neighbors (KNN), Bayesian ridge regression (BRR), logistic regression (LR), support vector machine (SVM), and decision tree (DT).  During the training phase, we run our models through a function to select the best hyperparameters using k-fold cross-validation, tuning each model based on its training data. We used these optimized models as our predictors for each experiment with each source dataset. As shown in Table~\ref{tbl:newResults}, our DT model achieved the highest R$^2$ score of 0.982, and RF achieved an R$^2$ of 0.981 after we added WI and MS data. KNN was a close third place with R$^2$ = 0.979. We detail the number of datapoints in each state for reference in Table~\ref{tbl:datapoints}. 

  

\section{Results}%

\begin{table}[!ht]
\centering
\begin{small}
\begin{tabular}{|c|c|c|}
\hline
    \textbf {Model}& \textbf {R$^2$} & \textbf {MAE (tons/acre)} \\ 
    \hline \hline
    DT & 0.928 & 0.091\\
    \hline
    RF & \textbf {0.941} & 0.081\\
    \hline
    KNN & 0.936 & 0.091\\
    \hline
    SVM & 0.917 & 0.094\\
    \hline
    BRR & 0.777 & 0.147\\
    \hline
    LR & 0.723 & 0.160\\
    \hline
\end{tabular}
\end{small}
\caption{\small Our best R$^2$ score overall, training on KY and GA data combined, was 0.941 with RF; RT, KNN, and SVM showed similar accuracies.}
\label{tbl:whitmireResults}
\vspace{-0.05in}
\end{table}

\begin{table}[!ht]
\centering
\begin{small}
\begin{tabular}{|c|c|c|}
\hline
    \textbf {Model}& \textbf {R$^2$} & \textbf {MAE (tons/acre)} \\ 
    \hline \hline
    DT & \textbf {0.982} & 0.240\\
    \hline
    RF & 0.981 & 0.247\\
    \hline
    KNN & 0.979 & 0.203\\
    \hline
    SVM & 0.900 & 0.392\\
    \hline
    BRR & 0.391 & 1.253\\
    \hline
    LR & 0.416 & 1.248\\
    \hline
\end{tabular}
\end{small}
\caption{\small Our best R$^2$ score overall after adding WI and MS increased to 0.982.}
\label{tbl:newResults}
\vspace{-0.05in}
\end{table}

\label{results}

In our first round of experiments, taking the traditional (non-DA) ML approach and training and testing on a combination of KY and GA data, we obtained our highest R$^2$ of 0.941 from the RF model, while our KNN and DT scores followed closely behind. In the second round, we repeated these experiments adding PA, WI, and MS data. While these results were also positive, we obtained better results when we removed PA, so we present the results from only KY, GA, MS, and WI in Table~\ref{tbl:newResults}, which shows our highest R$^2$ increased to 0.982 training on the expanded dataset, using the DT model. One interesting surprise was that we saw relatively high R$^2$ scores training on combinations that include WI, whose weather shares less in common with GA weather than MS or KY; WI experiences much lower winter temperatures and much lower soil moisture, among other differences. 

Next, we ran our DA-inspired ML experiments, where we use a trivial form of DA (TDA) in which we train on data from selected source states and test on data from one specific target state. As opposed to a true DA technique that mathematically augments the source data to make its distribution similar to the target data, TDA simply uses the data as-is. Overall, our TDA technique performed quite poorly, resulting in much lower accuracies than when we combine the states into one table, and both training and testing sets include multiple states. Though unsuccessful, TDA leaves plenty of room for improvement by using a true DA technique that augments the training data, and it shows us that simply training on specific source states and testing on one target state is not likely to yield good results in this domain.    

\begin{table}[!ht]
\centering
\begin{small}
\begin{tabular}{|c|c|}
\hline
    \textbf{State} & \textbf {No. of Datapoints}\\ 
    \hline \hline
    PA & 577\\
    \hline
    WI & 338\\
    \hline
    KY & 38\\
    \hline
    MS & 184\\
    \hline
    GA & 84\\
    \hline
\end{tabular}
\end{small}
\caption{\small Datasets per state with number of datapoints}
\label{tbl:datapoints}
\vspace{-0.05in}
\end{table}



\section{Conclusion}%
\label{conclusion}
We demonstrated that the problem of training ML models on scarce crop yield datasets for one geographical region can be partially remedied by training those models on data aggregated from other regions, even when they have disparate climates. We were able to improve on R{$^2$} scores obtained using traditional ML without DA, which is an important first step toward creating more sophisticated transfer learning and domain adaptation based pipelines, suggesting that further research into using these techniques is warranted.  

In the interest of improving machine learning's utility in the crucial field of agriculture, we expanded on previous work that compared feature selection techniques to pick the best features for predicting alfalfa crop yields. Toward the goal of creating a universal dataset that we can share broadly, we added virtually exhaustive data for three new U.S. states, for the timeframes available in variety trial reports. With more datapoints, we were able to interpret data on annual basis, instead of per harvest, which may be more meaningful.

Such data is widely collected in variety trials run by various universities but not in any uniform, organized format and not combined with any relevant weather data. This leads to artificial scarcity of data as different local regions' data prove challenging to combine. To deal with this scarcity, some previous works have implemented expensive, highly-specialized techniques for collecting data, such as using advanced remote sensors at specific locations. Such methods are typically not as widely adopted and do not cover comparable time spans to standard variety trials. Inspired by the ability of domain adaptation to reduce sparsity of data, we used our curated sources to learn models in order to make predictions on the target region (the original GA dataset). This technique yielded poor results, indicating that some true DA technique that augments data distributions should be explored and would probably yield better results. 

  

\section{Future Work}%
\label{future}
We are currently planning three main directions for our next projects with predicting alfalfa biomass yields. First, we plan to implement a variety of more sophisticated DA techniques than the one we present here and compare those results. Second, we plan to develop a time series strategy that helps us forecast future yields based preceding ones, adjusting its predictions along the way. Third, we are interested in analyzing how the different varieties perform in different geographical locations. We are also interested in streamlining our data curating pipeline to reduce the manual labor, and intend to introduce stabilization metrics into this future work.

  \section*{Acknowledgement}
  The authors would like to thank Sheng Li for his integral contributions to this work.
  
  \bibliographystyle{plainnat}
  \bibliography{ref.bib}
  
  \appendix

\section{Appendix}%
\label{appendix}

CODE AND DATA ACCESSIBILITY

The source code for this project is available at https://github.com/chriswhitmire/alfalfa-yield-prediction and https://github.com/thejonathanvancetrance/Alfalfa

The University of Georgia alfalfa yield data can be found here: https://georgiaforages.caes.uga.edu/species-and-varieties/cool-season/alfalfa.html

The University of Kentucky alfalfa yield data can be found as progress reports on this page: http://dept.ca.uky.edu/agc/pub\_prefix.asp?series=PR
Note that the only data that was used from the University of Kentucky was the non-roundup ready alfalfa varieties that were first harvested in the year 2013 or later.
The daily weather data for Kentucky, Pennsylvania, Mississippi, and Wisconsin was found on the National Oceanic and Atmospheric Administration website: https://www.ncdc.noaa.gov/crn/qcdatasets.html

The daily weather data for Georgia was given to us by the Georgia Automated Environmental Monitoring Network.

The day length was found from the United States Naval Observatory’s website: https://aa.usno.navy.mil/data/docs/Dur\_OneYear.php

Variety Trial Reports for PA can be found at https://extension.psu.edu/forage-variety-trials-reports

Variety Trial Reports for WI can be found at https://fyi.extension.wisc.edu/forage/category/trial-results/

Variety Trial Reports for MS can be found at https://www.mafes.msstate.edu/variety-trials/includes/forage/about.asp\#perennial


\section{Appendix}%
\label{appendixB}
HYPERPARAMETER GRID VALUES

The grid for the hyperparameters of each model is as follows:
Regression Tree:
\begin{itemize}
    \item ‘criterion': ['mae'],
    \item ‘max\_depth': [5,10,25,50,100]
\end{itemize}

Random forest:
\begin{itemize}
    \item 'n\_estimators': [5, 10, 25, 50, 100],
    \item 'max\_depth': [5, 10, 15, 20],
    \item 'criterion': ["mae"]
\end{itemize}

K-nearest neighbors:
\begin{itemize}
    \item 'n\_neighbors': [2,5,10],
    \item 'weights': ['uniform', 'distance'],
    \item 'leaf\_size': [5, 10, 30, 50]   
\end{itemize}
   
Support vector machine:
\begin{itemize}
    \item 'kernel': ['linear', 'poly', 'rbf', 'sigmoid'],
    \item 'C': [0.1, 1.0, 5.0, 10.0],
    \item 'gamma': ["scale", "auto"],
    \index 'degree': [2,3,4,5]
\end{itemize}
	 
Neural Network:
\begin{itemize}
    \item 'hidden\_layer\_sizes':[(3), (5), (10), (3,3), (5,5), (10,10)],
    \item 'solver': ['sgd', 'adam'],
    \item 'learning\_rate' : ['constant', 'invscaling', 'adaptive'],
    \index 'learning\_rate\_init': [0.1, 0.01, 0.001]    
\end{itemize}

Bayesian ridge regression:
\begin{itemize}
    \item 'n\_iter':[100,300,500],
    \item 'lambda\_1': [1.e-6, 1.e-4, 1.e-2, 1, 10],
    \item 'lambda\_1': [1.e-6, 1.e-4, 1.e-2, 1, 10]
\end{itemize}

Linear Regression: no hyperparameters
  
\end{document}